\documentclass[5p,authoryear]{elsarticle}
\usepackage{natbib}
\usepackage{amssymb}
\usepackage{graphicx}
\usepackage{amsmath}
\usepackage{amsfonts}
\usepackage{amsthm}
\usepackage{pifont}
\usepackage{txfonts}
\usepackage[colorlinks=true]{hyperref}

\journal{International Journal of Industrial Ergonomics}

\begin{document}

\begin{frontmatter}

\title{A novel approach for determining fatigue resistances of different muscle groups in static cases}

\author[TSINGHUA,ECN]{Liang MA\corref{cor}}
\ead{liangma@tsinghua.edu.cn; liang.ma@irccyn.ec-nantes.fr}
\author[IRCCyN]{Damien CHABLAT}
\author[IRCCyN,ECN]{Fouad BENNIS}
\author[TSINGHUA]{Wei ZHANG}
\author[TSINGHUA]{Bo HU}
\author[EADS]{Fran\c{c}ois GUILLAUME}
\address[TSINGHUA]{Department of Industrial Engineering, Tsinghua University, 100084, Beijing, P.R.China}
\address[IRCCyN]{ Institut de Recherche en Communications et en Cybern\'{e}tique de Nantes, UMR 6597  \\IRCCyN - 1, rue de la No\"{e}, BP 92 101 - 44321 Nantes CEDEX 03, France}
\address[ECN]{ \'{E}cole Centrale de Nantes, 1, rue de la No\"{e}, BP 92 101 - 44321 Nantes CEDEX 03, France}
\address[EADS]{EADS Innovation Works, 12, rue Pasteur - BP 76, 92152 Suresnes Cedex - France}

\cortext[cor]{Corresponding author: Tel:+86-10-62792665; Fax:+86-10-62794399}

\begin{abstract}
In ergonomics and biomechanics, muscle fatigue models based on maximum endurance time (MET) models are often used to integrate fatigue effect into ergonomic and biomechanical application. However, due to the empirical principle of those MET models, the disadvantages of this method are: 1) the MET models cannot reveal the muscle physiology background very well; 2) there is no general formation for those MET models to predict MET. 
In this paper, a theoretical MET model is extended from a simple muscle fatigue model with consideration of the external load and maximum voluntary contraction in passive static exertion cases. The universal availability of the extended MET model is analyzed in comparison to 24 existing empirical MET models. Using mathematical regression method, 21 of the 24 MET models have intraclass correlations over 0.9, which means the extended MET model could replace the existing MET models in a general and computationally efficient way. In addition, an important parameter, fatigability (or fatigue resistance) of different muscle groups, could be calculated via the mathematical regression approach. Its mean value and its standard deviation are useful for predicting MET values of a given population during static operations. The possible reasons influencing the fatigue resistance were classified and discussed, and it is still a very challenging work to find out the quantitative relationship between the fatigue resistance and the influencing factors.

\noindent{\textbf{Relevance to industry :}} 

MSD risks can be reduced by correct evaluation of static muscular work. Different muscle groups have different properties, and a generalized MET model is useful to simplify the fatigue analysis and fatigue modeling, especially for digital human techniques and virtual human simulation tools.
\end{abstract}


\end{frontmatter}

%% \linenumbers

%% main text
\section{Introduction}
\label{sec:intro}
Muscle fatigue is defined as ``any reduction in the ability to exert force in response to voluntary effort'' \citep{Chaffin1999}, and it is believed that the muscle fatigue is one of potential reasons leading to musculoskeletal disorders (MSDs) in the literature \citep{westgaard2000work,kumar2008musculoskeletal,kim2008knowledge}. Great effort has been contributed to integrate fatigue into different biomechanical models, especially in virtual human simulation for ergonomic application, in order to analyze the fatigue in muscles and joints and further to decrease  the MSD risks \citep{BADLER1999,Vignes2004,Hou2007}. 

In general, mainly two approaches have been adopted to represent muscle fatigue, either in theoretical methods or in empirical methods \citep{xia2008tam}. One or more decay terms were introduced into existing muscle force models in theoretical fatigue models, and those decay terms were mainly based on physiological performance of muscles in fatigue contraction. For example, a fatigue model based on the intracellular pH  \citep{Giat1993} was incorporated into Hill's muscle mechanical model \citep{Hill1938}. This fatigue model was also applied by \citet{KOMURA1999} to demonstrate the fatigue of different individual muscles. Another muscle fatigue model \citep{Wexler2003} based on physiological mechanism has been included into the Virtual Solider Research Program \citep{Vignes2004}, and in this model, dozens of parameters have to be fit for model identification only for a single muscle. As stated in \citet{xia2008tam}, ``these theoretical models are relatively complex but useful at the single muscle level. However, they do not readily handle task-related biomechanical factors such as joint angle and velocity.'' Meanwhile, several muscles around a joint are engaged in order to realize an action or a movement around the joint, and mathematically this results in an underdetermined equation while determining the force of each engaged muscle due to muscle redundancy and complex muscle force moment arm-joint angle relationships. Although different optimization methods have been used to face this load sharing problem \citep{ren2007pmh}, it is still very difficult to validate the optimization result and further the fatigue effect, due to the complexity of anatomical structure and the physiological coordination mechanism of the muscles. 

Muscle fatigue is often modeled and extended based on Maximum Endurance Time (MET) models at joint level in empirical methods. These models are often used in ergonomic applications to handle task-related external parameters, such as intensity of the external load, frequency of the external load, duration, and posture. In these models, the MET of a muscle group around a joint was often measured under static contraction conditions until exhaustion. Using this method can avoid complex modeling of individual muscles, and net joint strengths already exist in the literature for determining the relative load \citep{anderson2007mvj, xia2008tam}. The most famous one of these MET models is  Rohmert's curve \citep{rohmert1960ees} which was usually used as guideline for designing the static contraction task. Besides Rohmert's MET model, there are several other empirical MET models in the literature \citep{Khalid2006}. These MET models are very useful to evaluate physical fatigue in static operations and to determine work-rest allowances  \citep{el2009comparison}, and they were often employed in biomechanical models in order to minimize fatigue as well. For example, \citet{Freund2001} proposed a dynamic model for forearm in which the fatigue component was modeled for each single muscle by fitting Rohmert's curve in \citet{Chaffin1999}. \citet{Rodriguez2002} proposed a half-joint fatigue model, more exactly a fatigue index, based on mechanical properties of muscle groups. The holding time over maximum endurance time is used as an indicator to evaluate joint fatigue. In \citet{niemi1996ssm}, different fiber type composition was taken into account with endurance model to locate the muscle fatigue into single muscle level. However, in MET models, the main limitations are: 1) The physical relationship in these models cannot be interpreted directly by muscle physiology, and there is no universality among these models. 2) All the MET models were achieved by fitting experimental results using different formulation of equation. It has been found that muscle fatigability can vary across muscles and joints. However, there is no general formulation for those models. 3) Differences have been found among those MET models for different muscle groups, for different postures, and even for different models for the same muscle group. Due to the limitation from the empirical principle, the differences cannot be interpreted by those MET models. Thus, it is necessary to develop a general MET model which is able to replace all the experimental MET models and explain all the differences cross these models.

\citet{xia2008tam} proposed a new muscle fatigue model based on motor units (MU) recruitment \citep{Jing2002} to combine the theoretical models and the task-related muscle fatigue factors. In this model, properties of different muscle fiber types have been assumed to predict the muscle fatigue at joint level. However, in their research, the validation of their fatigue model was not provided. Furthermore, the different fatigability of different muscle groups has not been analyzed in details in this model. Fatigability (the reciprocal of endurance capacity or the reciprocal of fatigue resistance) can be defined by the endurance time or measured by the number of times of an operation until exhaustion. This measure is an important parameter to measure physical fatigue process during manual handling operations \citep{lynch1999mqa, clark2003ddp, lariviere2006gif, hunter2004ias}.

In \citet{ma2008nsd}, we constructed a new muscle fatigue model in which the external task related parameters are taken into consideration to describe physical fatigue process, and this model has also been interpreted by the physiological mechanism of muscle. The model has been compared to 24 existing MET models, and great linear relationships have been found between our model and the other MET models. Meanwhile, this model has also been validated in comparison to three theoretical models. This model is a simpler, theoretical approach to describe the fatigue process, especially in static contraction cases.

In this paper, further analysis based on the fatigue model is carried out using mathematical regression method to determine the fatigability of different muscle groups. We are going to propose a mathematical parameter, defined as fatigability, describing the resistance to the decrease of the muscle capacity. The fatigue resistance for different muscle groups is going to be regressed from experimental MET models. The theoretical approach for calculating the fatigue resistance will be explained in Section \ref{sec:method}. The muscle fatigue model in \citet{ma2008nsd} is going to be presented briefly in Section \ref{sec:model}. A general MET model is extended from this fatigue model in Section \ref{sec:met}. The mathematical procedure for calculating the fatigability contributes to the main content of Section \ref{sec:regression}. The results and discussion are given in Section \ref{sec:result} and \ref{sec:discussion}, respectively.

\section{Method} \label{sec:method}
\subsection{A dynamic muscle fatigue model} \label{sec:model}
A dynamic fatigue model based on muscle active motor principle was proposed in \citet{ma2008nsd}. This model was able to integrate task parameters (load) and temporal parameters into manual handling operation in industry. The differential equation for describing the reduction of the capacity is Eq. (\ref{eq:FcemDiff}). The descriptions of the parameters for Eq. (\ref{eq:FcemDiff}) are listed in Table \ref{tab:Parameters}. 

\begin{equation}
\label{eq:FcemDiff}
			\frac{dF_{cem}(t)}{dt} = -k \frac{F_{cem}(t)}{MVC}F_{load}(t)
\end{equation}

\begin{table*}[htbp]
	\centering
	\caption{Parameters in Dynamic Fatigue Model}
	\label{tab:Parameters}
		\begin{tabular}{lcl}
		\hline
		Item & Unit & Description\\
		\hline
		$MVC$					& $N$ &	Maximum voluntary contraction, maximum capacity of muscle\\
		$F_{cem}(t)$ 	& $N$ & Current exertable maximum force, current capacity of muscle\\
		$F_{load}(t)$	& $N$ & External load of muscle, the force which the muscle needs to generate\\
		$k$						& $min^{-1}$ & Constant value, fatigue ratio, here $k=1$\\
		$\%MVC$				&				&Percentage of the voluntary maximum contraction\\
		$f_{MVC}$			&				&$\%MVC/100$\\
		\hline			
		\end{tabular}
\end{table*}

The fatigue model in Eq. (\ref{eq:FcemDiff}) can be explained by the MU-based pattern of muscle \citep{Jing2002,Vollestad1997}. According to muscle physiology,  larger $F_{load}$ means more type II motor units are involved into the force generation. As a result, the muscle becomes fatigued more rapidly, as expressed in Eq. (\ref{eq:FcemDiff}). $F_{cem}$ represents the non-fatigue motor units of the muscle. The decreased fatigability is expressed by term $F_{cem}(t)/MVC$ due to the composition change in the non-fatigue muscle fibers. $k$ is a parameter indicating the fatigue ratio, and it has different constant values for different muscle groups. In a general case, it is assigned as 1. This dynamic model has been mathematical validated in \citet{ma2008nsd} with static MET models and other existing dynamic theoretical models. Much more detailed explanation can be found in \citet{ma2008nsd}.

The integration result of Eq. (\ref{eq:FcemDiff}) is Eq. (\ref{eq:FcemInt}), if $F_{cem}(0)= MVC$. Equation (\ref{eq:FcemInt}) demonstrates the fatigue process under an external load $F_{load}(u)$. The external load can be either static external load or dynamic one.
\begin{equation}
\label{eq:FcemInt}
			F_{cem}(t) = MVC \, e^{\int_{0}^{t} -k \dfrac{F_{load}(u)}{MVC}du}
\end{equation}

\subsection{The extended MET model} \label{sec:met}

A general MET model can be extended by supposing that $F_{load}(t)$ is constant in static situation to predict the endurance time of a muscle contraction. $MET$ is the duration in which $F_{cem}$ falls down to the current $F_{load}$. Thus, $MET$ can be determined by Eq. (\ref{eq:DynamicEndurance1}) and (\ref{eq:DynamicEndurance2}).

\begin{equation}
\label{eq:DynamicEndurance1}
	F_{cem}(t) = MVC \, e^{\int_{0}^{t} -k \dfrac{F_{load}(u)}{MVC}du} = F_{load}(t)
\end{equation}

\begin{equation}
\label{eq:DynamicEndurance2}
			t = MET = -\dfrac{ ln\left({\dfrac{F_{load}(t)}{MVC}}\right)}{k\dfrac{F_{load}(t)}{MVC}} =-\dfrac{ ln(f_{MVC})}{k\, f_{MVC}}
\end{equation}

In comparison to the empirical MET models, the extended MET model is obtained from a theoretical muscle fatigue model based on motor units recruitment mechanism. In the extended MET model, only one parameter $k$ remains variable. This parameter is defined as fatigability (fatigue ratio) in the theoretical model, and this parameter could enable us to explain the differences found in empirical MET models. 

\subsection{Regression for determining the fatigability}
\label{sec:regression}
In \citet{ma2008nsd}, two correlation coefficients were selected to analyze the relationship between the extended MET model ($k=1$) and the empirical MET models. One is Pearson's correlation $r$ in for evaluating the linear relationship and the other one is intraclass correlation ($ICC$) for evaluating the similarity between two models. The calculation results are shown in columns $r$ and $ICC_1$ of Table \ref{tab:StaticValidation}. There are still large differences from 1 in $ICC_1$ column, while high linear correlations have also been found between the extended MET model and the empirical MET models. That means that the extended MET model can fit the empirical MET models for different muscle groups by adjusting the parameter $k$. 

A mathmatical regression process is carried out to determine the parameter $k$ as follows. From Table \ref{tab:StaticValidation}, it is observed that almost all the static MET models have high linear relationship with the extended MET model (for most models, $r>0.95$), which means that each static model can be described mathematically by a linear equation (Eq. (\ref{eq:linear})). In Eq. (\ref{eq:linear}), $x$ is used to replace $f_{MVC}$ and $p(x)$ represents the extended MET model in Eq. (\ref{eq:DynamicEndurance2}). $\;m$ and $n$ are constants describing the linear relationship between an existing MET model and the extended MET model, and they are needed to be determined in linear regression. It should be noticed that $m=1/k$ indicates the fatigue resistance of the static model, and $k$ is fatigue ratio or fatigability of different static model.

\begin{equation}
\label{eq:linear}
	f(x)=m\,p(x)+n
\end{equation}

Due to the asymptotic tendencies of the empirical MET models mentioned in \citet{Khalid2006}, when $x \to 1$ ($\%MVC \to 100$), $f(x) \to 0$ and $p(x) \to 0$ ($MET \to 0$), therefore, we can assume that $n=0$. For this reason, only one parameter $m$ needs to be determined. Since some empirical MET models are not available for $\%MVC$ under 15\%, the regression was carried out in the interval from $x=0.16$ to $x=0.99$. With an space 0.01, $N=84$ empirical MET values were calculated to determine the parameter $m$ by minimizing the function in Eq. (\ref{eq:objective}).

\begin{equation}
\label{eq:objective}
	M(x)=\sum\limits_{i=1}^{N}\left(f(x_i)-m\,p(x_i)\right)^2=a\,m^2+b\,m+c
\end{equation}

In Eq. (\ref{eq:objective}), $a=\sum\limits_{i=1}^{N}p(x_i)^2$ is always greater than 0, and $b=-2\sum\limits_{i=1}^{N}p(x_i)f(x_i)$ is always less than 0, therefore the fatigue resistance $m$  can be calculated by Eq. (\ref{eq:regression}).
\begin{equation}
\label{eq:regression}
	m=\dfrac{-b}{2a}=\dfrac{\sum\limits_{i=1}^{N}p(x_i)f(x_i)}{\sum\limits_{i=1}^{N}p(x_i)^2}\,>0
\end{equation}

After regression for each empirical MET model, new $ICC$ values were calculated by comparing $f(x)/m$ and $p(x)$, and they are listed in the column $ICC_2$ of Table \ref{tab:StaticValidation}. It is easy to prove that the regression does not change the $r$ correlation. For this reason, only $ICC$ is recalculated.

\section{Results} \label{sec:result}
\subsection{Regression results}
Both $ICC$ correlations before regression and after regression are shown in Table \ref{tab:StaticValidation}. It should be noticed that the results $ICC_1$ before regression in Table \ref{tab:StaticValidation} were slightly different from the results presented in \citet{ma2008nsd}, because the range of $f_{MVC}$ varies from 0.15 to 0.99 in this paper while it varied from 0.20 to 0.99 in \citet{ma2008nsd} in order to validate the dynamic fatigue model. Some models were sensitive for such a change, e.g. Monod's model. However, the little change of the validation result does not change the conclusion in \citet{ma2008nsd}. 

\begin{figure*}[hbt]
   \begin{center}
   \begin{tabular}{cc}
        \begin{minipage}[t]{0.45\textwidth}
				\centering
				\includegraphics[width=\textwidth]{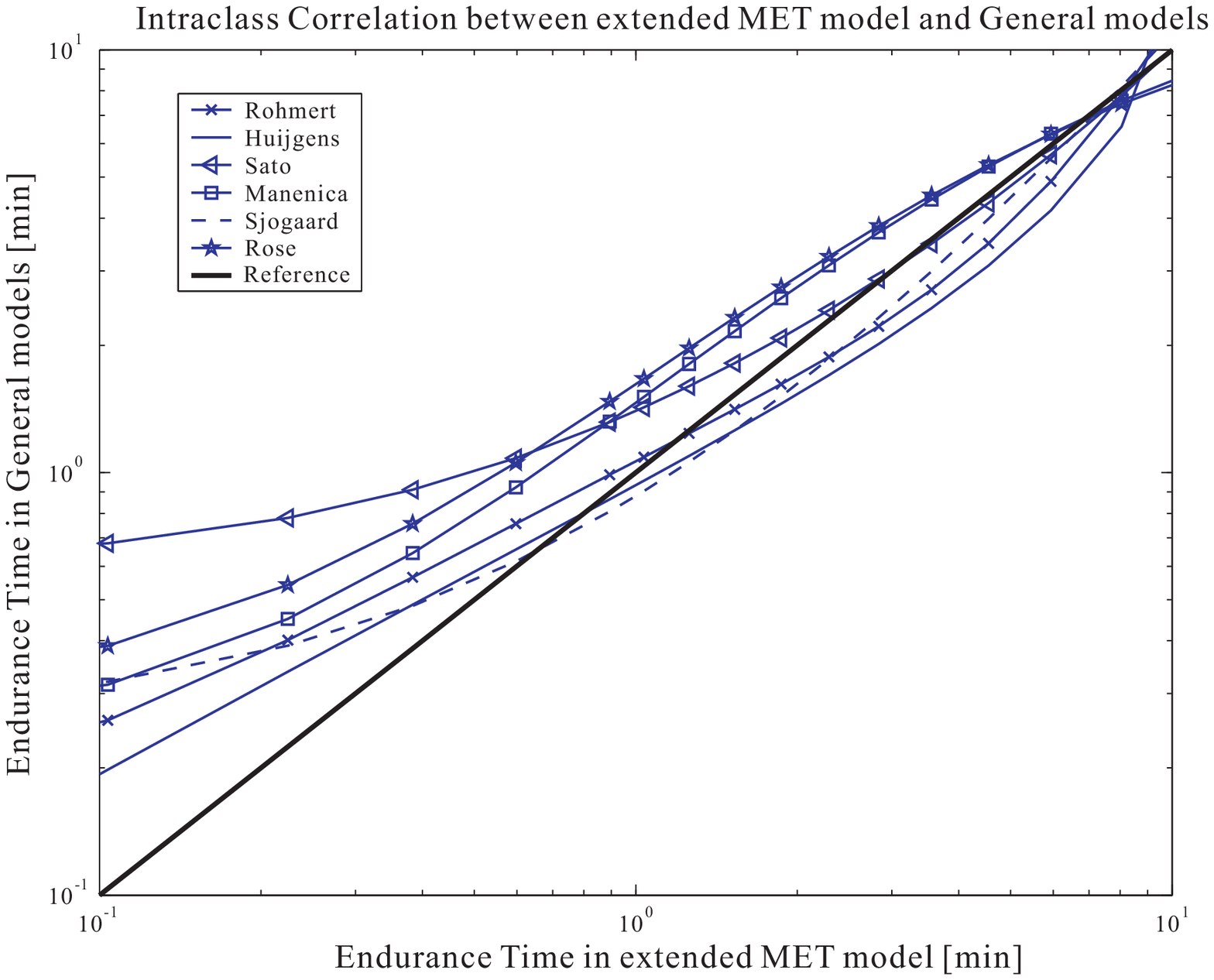}
				\caption{ICC diagram for MET general models after regression}
				\label{fig:generalafter}
	    \end{minipage} &
		\begin{minipage}[t]{0.45\textwidth}
				\centering
				\includegraphics[width=\textwidth]{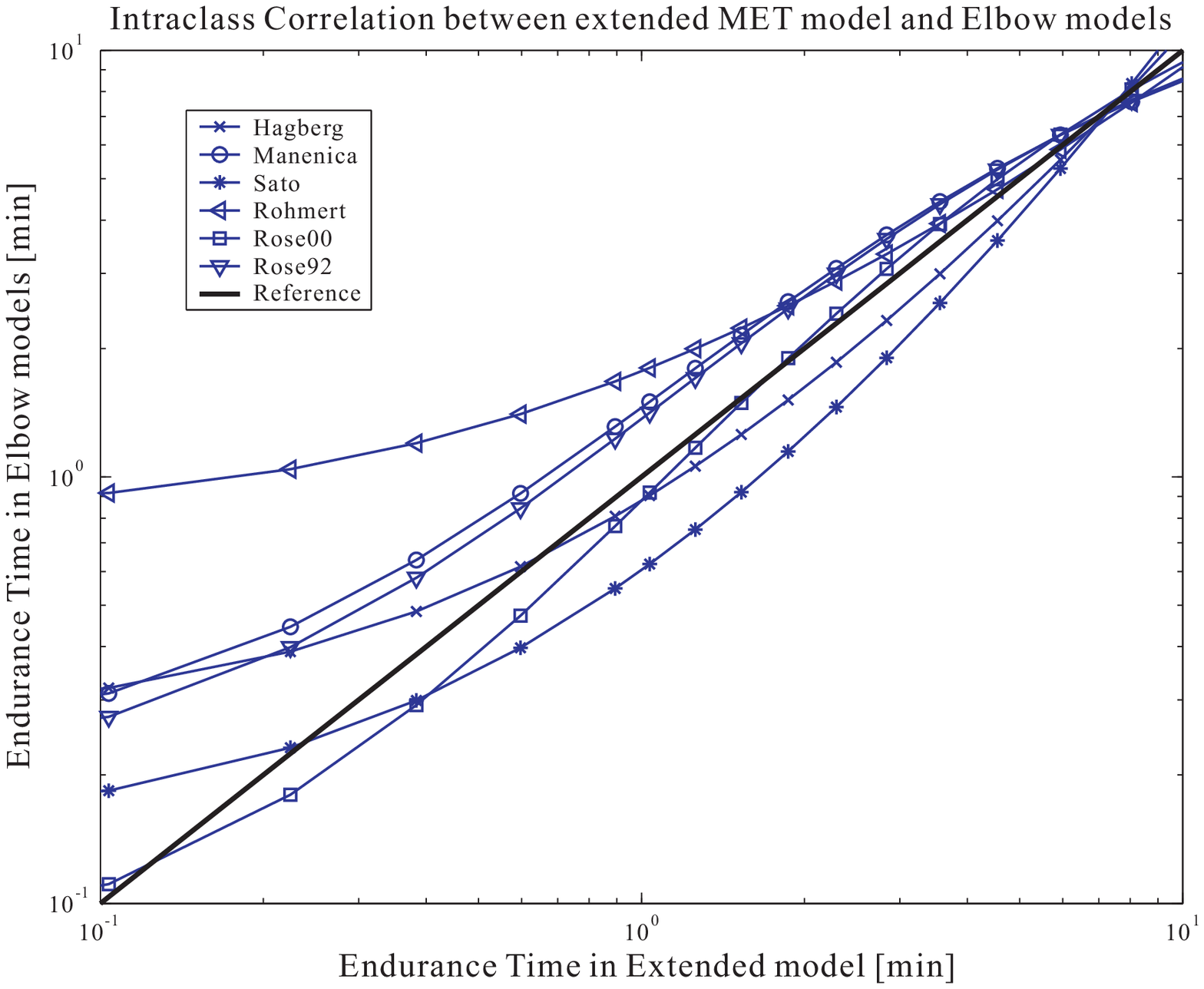}
				\caption{ICC diagram for MET elbow models after regression}
				\label{fig:elbowafter}
	    \end{minipage} 
    \end{tabular}
   \end{center}
\end{figure*}

\begin{figure*}[hbt]
   \begin{center}
   \begin{tabular}{cc}

      \begin{minipage}[t]{0.45\textwidth}
				\centering
				\includegraphics[width=\textwidth]{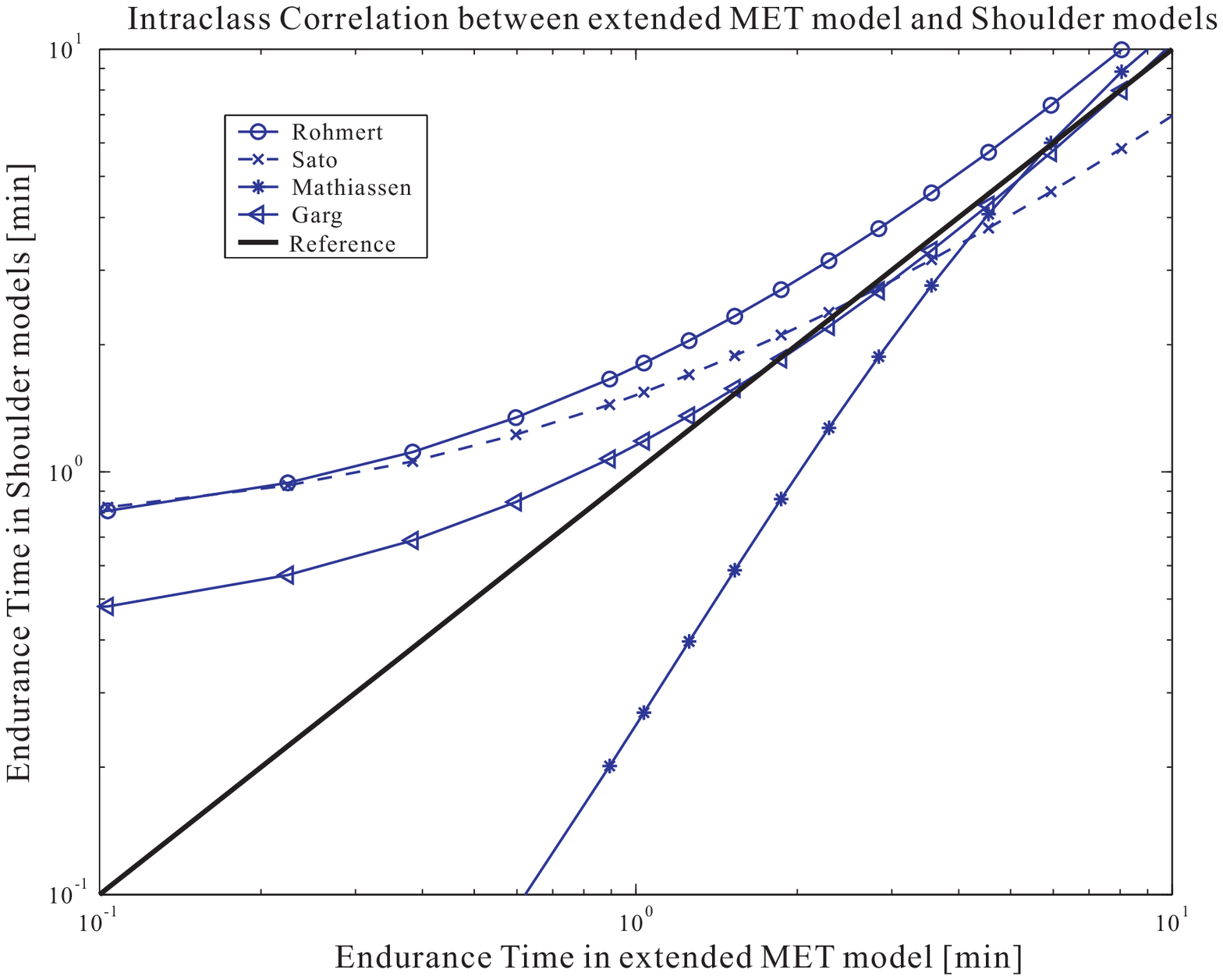}
				\caption{ICC diagram for MET shoulder models after regression}
				\label{fig:shoulderafter}
	    \end{minipage} &
		      \begin{minipage}[t]{0.45\textwidth}
				\centering
				\includegraphics[width=\textwidth]{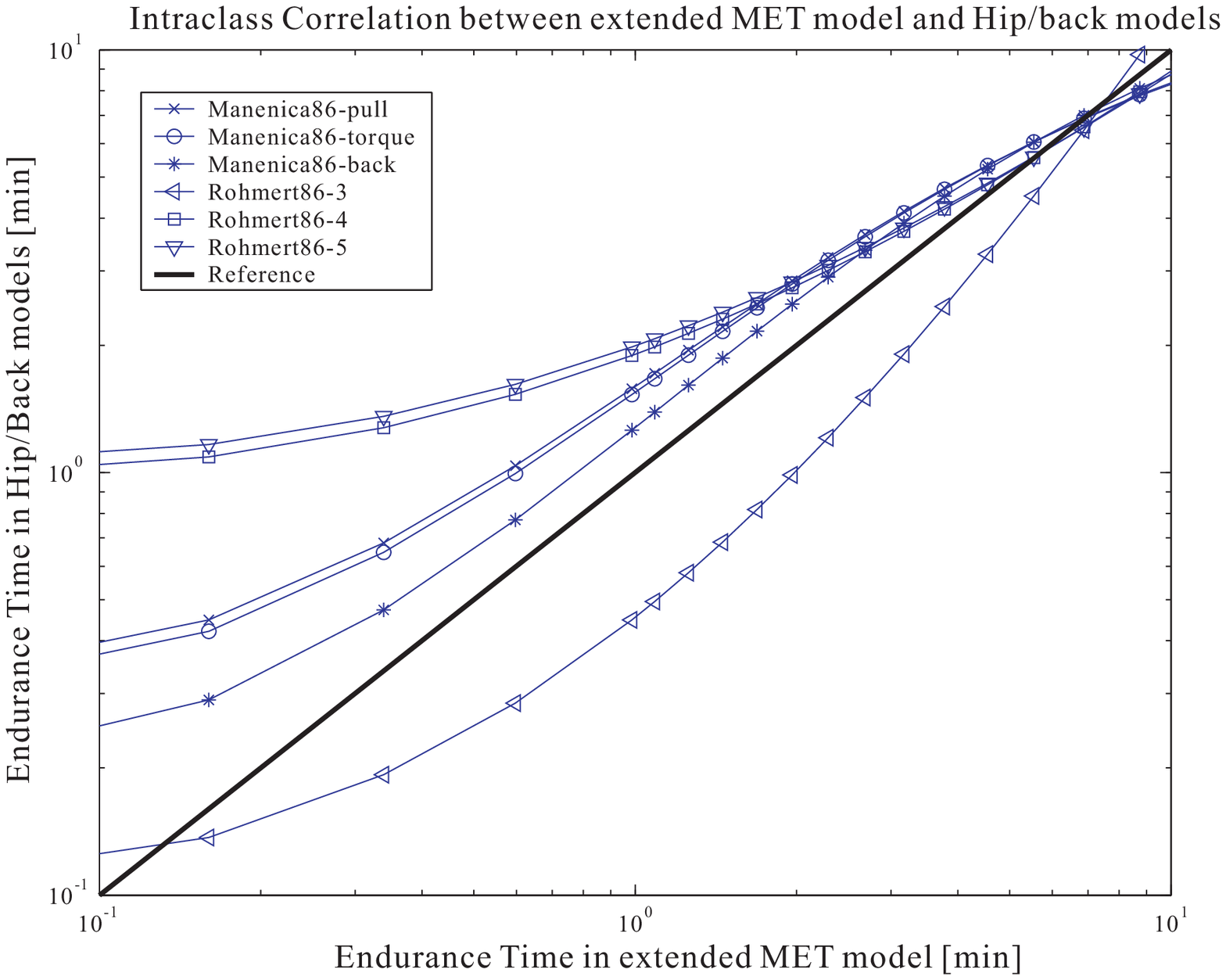}
				\caption{ICC diagram for Hip/back shoulder models after regression}
				\label{fig:hipafter}
	    \end{minipage} 
    \end{tabular}
   \end{center}
\end{figure*}

The $ICC$ results for different muscle groups after regression are graphically presented in log-log diagram from Fig. \ref{fig:generalafter} to \ref{fig:hipafter} as well. The corresponding graphical results before regression can be found in \citet{ma2008nsd}. The x-axis is the MET values predicted using the extended MET model at different relative force levels, while the y-axis is the MET values predicted using other MET models at corresponding relative force levels. The straight solid diagonal line is the reference line. For the other models, the one which approaches more closely to the straight line has a higher $ICC$. If a model coincides completely over the reference line, which means it is identical to the extended MET model. 

\begin{table*}[htbp]
	\centering
	\caption[Static validation results]{Static validation results  $r$ and $ICC$ between the extended MET model and the other existing $MET$ models in \protect \citet{Khalid2006}}
	\label{tab:StaticValidation}
		\begin{tabular}{llccc}
		\hline
		Model & MET equations (in minutes) & r & $ICC_1$ &$ICC_2$\\
		\hline
		\textbf{\textit{General models}}\\
		\hline
		Rohmert& $MET=-1.5+\frac{2.1}{f_{MVC}}-\frac{0.6}{f_{MVC}^2} +\frac{0.1}{f_{MVC}^3}$&0.9717\; &0..9505\;&0.9707\\
		Monod and Scherrer& $MET=0.4167\,(f_{MVC}-0.14)^{-2.4}$&0.6241\;  &0.0465\;&0.4736\\
		Huijgens& $MET=0.865\, \left[{\frac{1-f_{MVC}}{f_{MVC}-0.15}}\right]^{1/1.4}$&0.9036\; &0.8947\;&0.8916\\
		Sato et al.& $MET=0.3802 \, (f_{MVC}-0.04)^{-1.44}$&0.9973\; &0.8765\;&0.9864\\
		Manenica& $MET=14.88\, \exp(-4.48f_{MVC})$&0.9829\; &0.9357\;&0.9701\\
		Sjogaard& $MET=0.2997\, f_{MVC}^{-2.14}$&0.9902\; &0.9739\;&0.9898\\
		Rose et al.& $MET=7.96\, \exp(-4.16f_{MVC})$&0.9783\; &0.6100\;&0.9573\\
		\hline
		\textbf{\textit{Upper limbs models}}\\
		\hline
		\textit{Shoulder}\\
		Sato et al.& $MET=0.398\, f_{MVC}^{-1.29}$&0.9988\; &0.5317\;&0.9349\\
		Rohmert et al.& $MET=0.2955\, f_{MVC}^{-1.658}$&0.9993\; &0.7358\;&0.8982\\
		Mathiassen and Ahsberg& $MET= 40.6092\, \exp(-9.7f_{MVC})$&0.9881\; &0.8673\;&0.9711\\
		Garg& $MET= 0.5618 \, f_{MVC}^{-1.7551}$&0.9968\; &0.9064\;&0.9947\\
		\hline
		\textit{Elbow}\\
		Hagberg& $MET=0.298\, f_{MVC}^{-2.14}$&0.9902\; &0.9751\;&0.9898\\
		Manenica& $MET=20.6972\, \exp(-4.5f_{MVC})$&0.9832\; &0.9582\;&0.9708\\
		Sato et al.& $MET=0.195\, f_{MVC}^{-2.52}$&0.9838\; &0.9008\;&0.9688\\
		Rohmert et al.& $MET=0.2285\, f_{MVC}^{-1.391}$&0.9997\; &0.2942\;&0.9570\\
		Rose et al.2000& $MET=20.6\, \exp(-6.04f_{MVC})$&0.9958\; &0.9627\;&0.9708\\
		Rose et al.1992& $MET=10.23\, \exp(-4.69f_{MVC})$&0.9855\; &0.7053\;&0.9766\\
		\hline
		\textit{Hand}\\
		Manenica& $MET=16.6099\, \exp(-4.5f_{MVC})$&0.9832\; &0.9840\;&0.9646\\
		\hline
		\textbf{\textit{Back/hip models}}\\
		\hline
		Manenica (body pull)& $MET=27.6604\, \exp(-4.2f_{MVC})$&0.9789\; &0.7672\;&0.9591\\
		Manenica (body torque)& $MET=12.4286\, \exp(-4.3f_{MVC})$&0.9804\; &0.8736\;&0.9634\\
		Manenica (back muscles)& $MET=32.7859\, \exp(-4.9f_{MVC})$&0.9878\; &0.8091\;&0.9819\\
		Rohmert (posture 3)& $MET=0.3001\, f_{MVC}^{-2.803}$&09655\; &0.4056\;&0.9482\\
		Rohmert (posture 4)& $MET=1.2301\, f_{MVC}^{-1.308}$&0.9990\; &0.8356\;&0.9396\\
		Rohmert (posture 5)& $MET=3.2613\, f_{MVC}^{-1.256}$&0.9984\; &0.1253\;&0.9263\\
		\hline							
		\end{tabular}
\end{table*}

\subsection{Fatigue resistance results}

The regression results ($m$) for each MET model are listed in Table \ref{tab:FatigueResistanceTable}. The mean value $\bar{m}$ and the standard deviation $\sigma_m$ were calculated for different muscle groups as well. The Monod's general model is eliminated from the calculation due to its poor $ICC$ value. The intergroup differences are represented by the mean value of each muscle group. The Hip/back models have a higher mean value $\bar{m}=1.9701$, while the other human body segments and the general models have relative lower fatigue resistances ranging from 0.76 to 0.90, without big differences. The fluctuation in each muscle group, namely the intra muscle group difference, is presented by $\sigma_m$. The stability in the general group is the best, and the hip/back model has the largest variation. There is no big difference between elbow and shoulder models. 

\begin{table*}[htbp]
	\centering
	\caption{Fatigue resistance $m$ of different MET models}
		\begin{tabular}{cccccccccc}
		\hline
		Segment&$m_1$&$m_2$&$m_3$&$m_4$&$m_5$&$m_6$&$m_7$&$\bar{m}$&$\sigma_m$\\
		\hline
		General&Rohm.\;&Mono.\;&Hijg.\;&Sato.\;&Mane.\;&Sjog.&Rose\;\\
		%&0.8392\;&\bf{0.399}\;&0.855\;&0.8346\;&1.0785\;&1.1114\;&0.6465\;&0.8942\;&0.030\\
		&0.8328\;&-\;&0.9514\;&0.6836\;&0.8019\;&1.1468\;&0.4647\;& 0.8135\;& 0.2320\\

		\hline
		Shoulder&Sato.\;&Rohme.\;&Math.\;&Garg\\
		&0.4274\;&0.545\;&0.698\;&1.3926\;&\;&\;&\;&0.7562\;&0.4347\\
		\hline
		Elbow&Hagb.\;&Mane.\;&Sato.\;&Rohm.\;&Rose00\;&Rose92\;\\
		&1.1403\;&1.1099\;&1.3461\;&0.2842\;&0.7616\;&0.5234\;&\;&0.8609\;&0.4079\\
		\hline
		Hand&Mane.\\
		&0.8907\;&&&&&&&0.8907&-\\
		\hline
		Hip&pull\;&torq.\;&back\;&pos3\;&pos4\;&pos5\;\\
		&1.5986\;&0.7005\;&1.5931\;&3.2379\;&1.356\;&3.3345\;&\;&1.9701\;&1.1476\\
		\hline			
		\end{tabular}
	\label{tab:FatigueResistanceTable}
\end{table*}

\section{Discussion} 
\label{sec:discussion}

\subsection{Result analysis}
\textbf{ICC}: Almost all the $ICC_2$ are greater than 0.89, and only one is an exception (Monond and Scherrer, 0.4736). This exception is because of its relative worse linear correlation $r$ with the extended MET model, while almost all the other ones have $r$ over 0.96, and the Monod's model has only 0.6241. For the Monod's model, the linear error occurs mainly when the $f_{MVC}$ approaches to 0.15. This error is mainly caused by the way in which the Monod's model is formulated. This exception is eliminated in the following analysis and discussion.

There are larger differences between the extended MET model and the empirical MET models, especially when the $f_{MVC}$ approaches to 0.15. Those differences can be explained by the interindividual difference in MET, and these differences are greater for the low \%MVC \citep{Khalid2006}. From the graphical representation, it can be noticed that the MET errors are mainly decreased in the range from $10^0\,min$ to $10^1\,min$, which means the extended MET model after regression can predict MET with less error than using the extended MET model before regression(refer to Fig.2, Fig. 4, Fig. 6 and Fig. 8 in \citet{ma2008nsd}). 

The greatest improvement of the fitness between the extended MET model and the empirical MET models is the Hip/Back model (Fig. \ref{fig:hipafter}). This approves that the extended MET model with a suitable fatigue ratio can adapt itself well to the most complex part of human body. The same improvement can be found for shoulder models and most of the elbow models. It should be noticed not all the MET models have been improved after the regression. Little fall can be found for the MET models (hand model) with $ICC$ over 0.98 in the $ICC_1$ column. The possible reason is that it has already relative high $ICC$ correlation, and the regression does not improve its fitness. However, those models after regression still have high $ICC$ ($>0.95$). As a summary, the regression approach achieves high $ICC$ and improves the similarity between the extended MET model and the existing models. This proves that the extended MET model can be adapted to fit different body parts, and the extended MET model can predict the MET for static cases.

\begin{figure*}[hbt]
   \begin{center}
   \begin{tabular}{cc}
    \begin{minipage}[t]{0.45\textwidth}
       	\centering
				\includegraphics[width=1\textwidth]{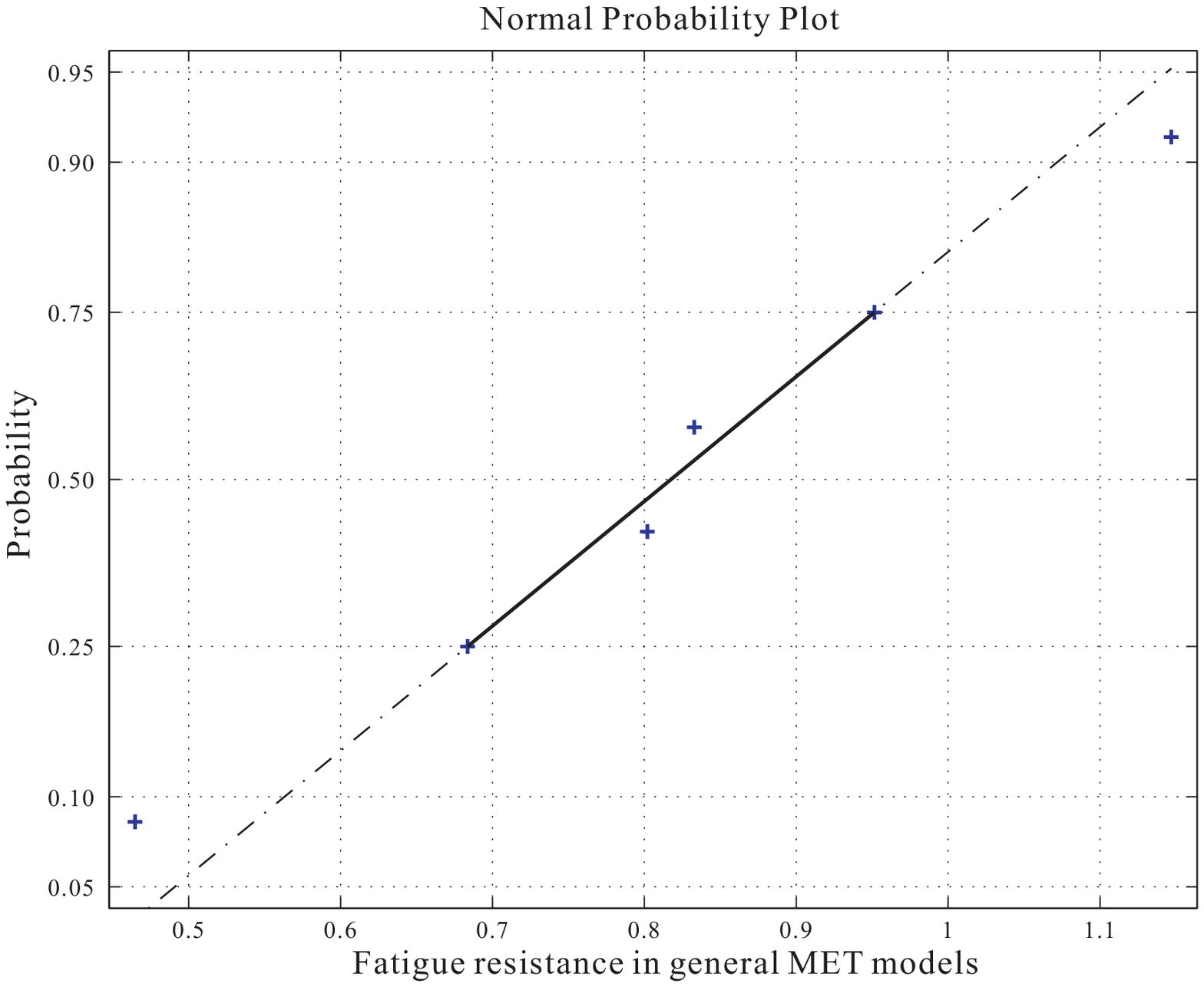}
				\caption{Normal distribution test for the general model}
				\label{fig:normalgeneral}
      \end{minipage}&
      \begin{minipage}[t]{0.45\textwidth}
				\centering
				\includegraphics[width=\textwidth]{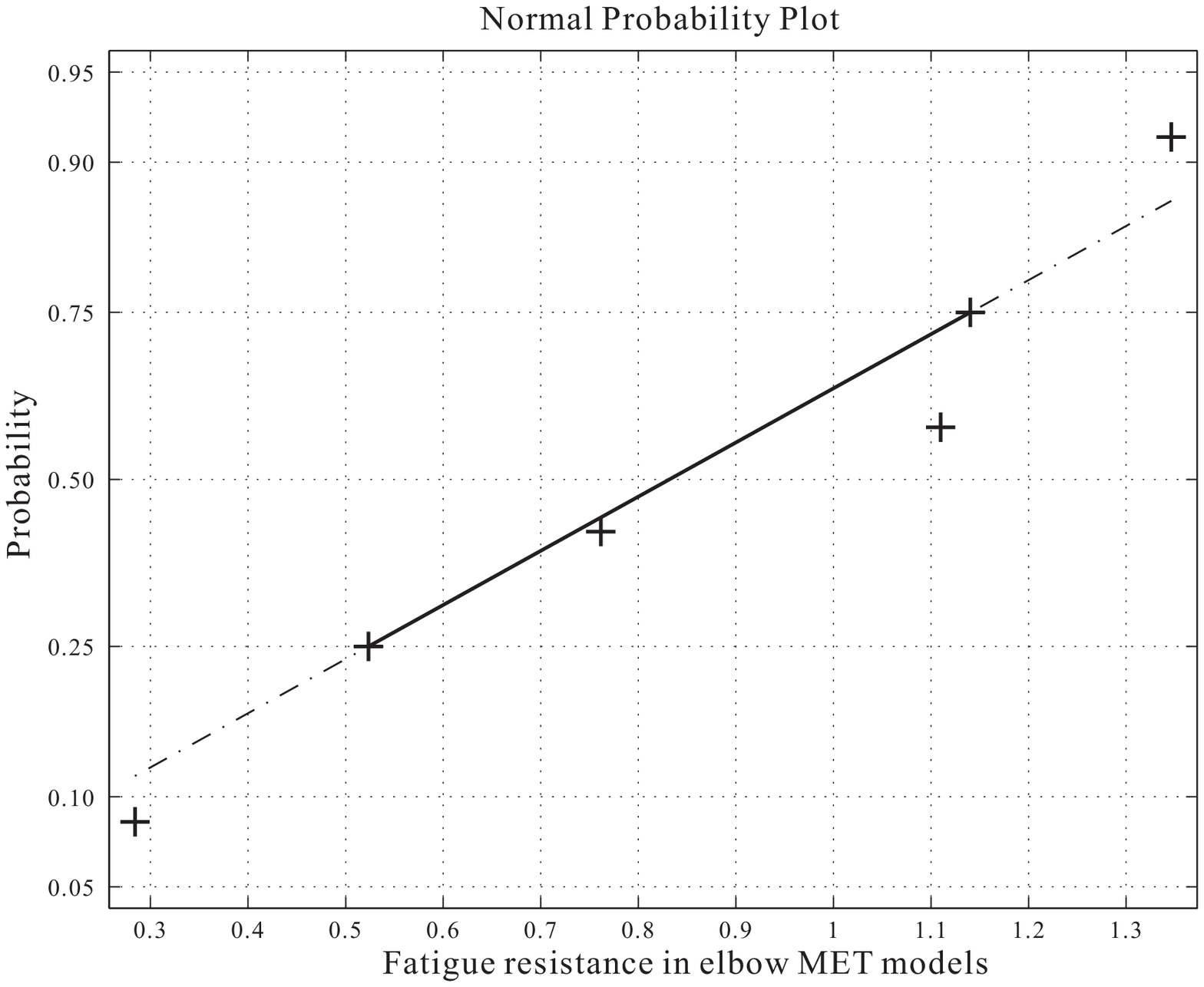}
				\caption{Normal distribution test for the elbow model}
				\label{fig:normalelbow}
	    \end{minipage} 
    \end{tabular}
   \end{center}
\end{figure*}

\textbf{Fatigue resistance}: The regression result of the fatigue resistance of different muscle groups were tested with normplot function in Matlab in order to graphically assess whether the fatigue resistances could come from a normal distribution. The test result shows fatigue resistances for general models and elbow models scatter near the diagonal line in the Fig. \ref{fig:normalgeneral} and Fig. \ref{fig:normalelbow}. Due to limitation of sample numbers in shoulder models and the large variance in hip/back models, the distribution test did not achieve satisfying result. Once there are enough sample models, it can be extrapolated that the fatigue resistances for different muscle groups for the overall population distributes in normal probability, therefore, the mean value locates in $\bar{m}\pm\sigma$ could predict the fatigue property of 50\% population.

\begin{figure*}[hbt]
   \begin{center}
   \begin{tabular}{cc}
    \begin{minipage}[t]{0.45\textwidth}
       	\centering
				\includegraphics[width=1\textwidth]{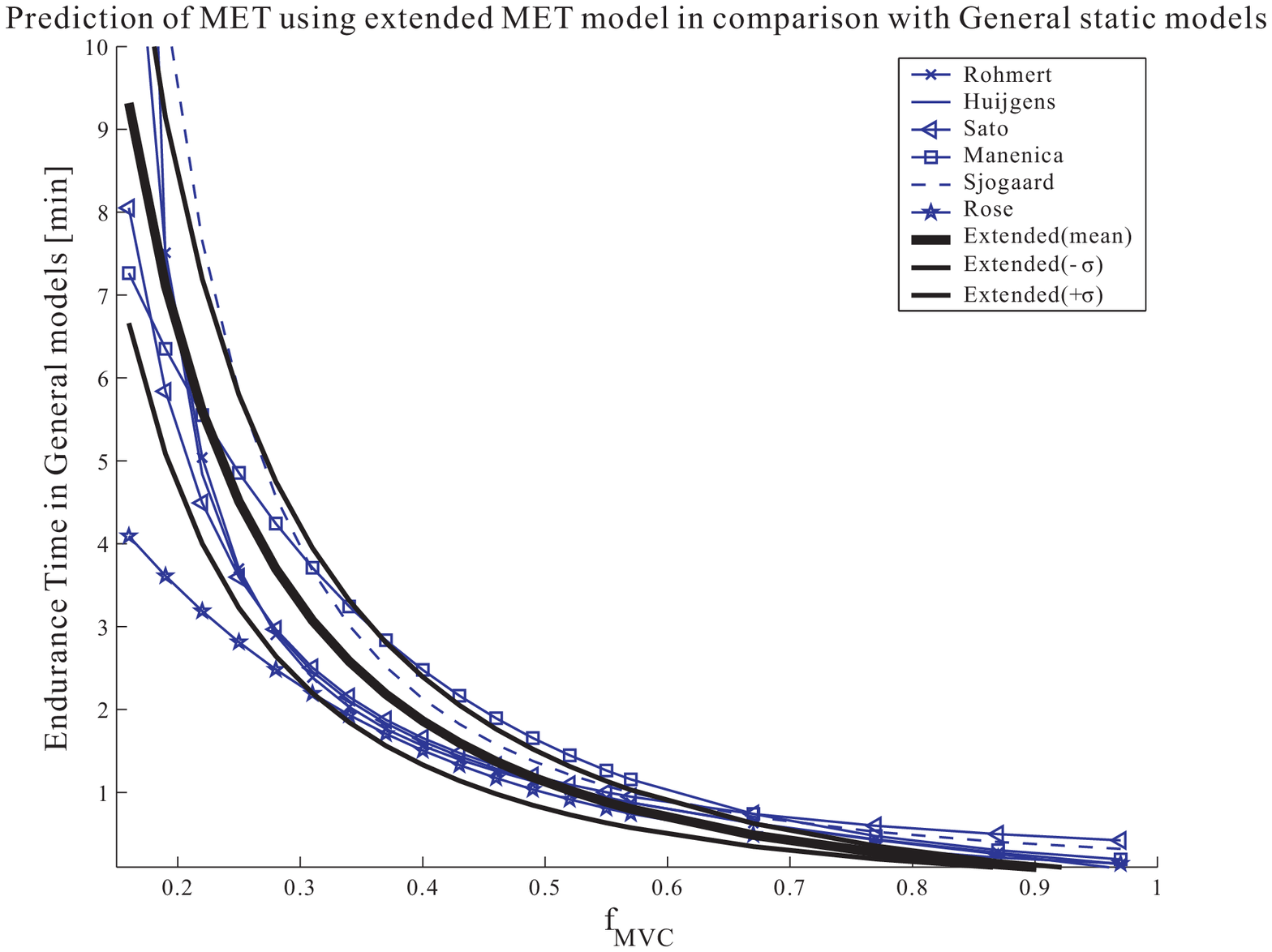}
				\caption{MET prediction using extended MET model in comparison with general static models}
				\label{fig:generalresult}
      \end{minipage}&
      \begin{minipage}[t]{0.45\textwidth}
				\centering
				\includegraphics[width=\textwidth]{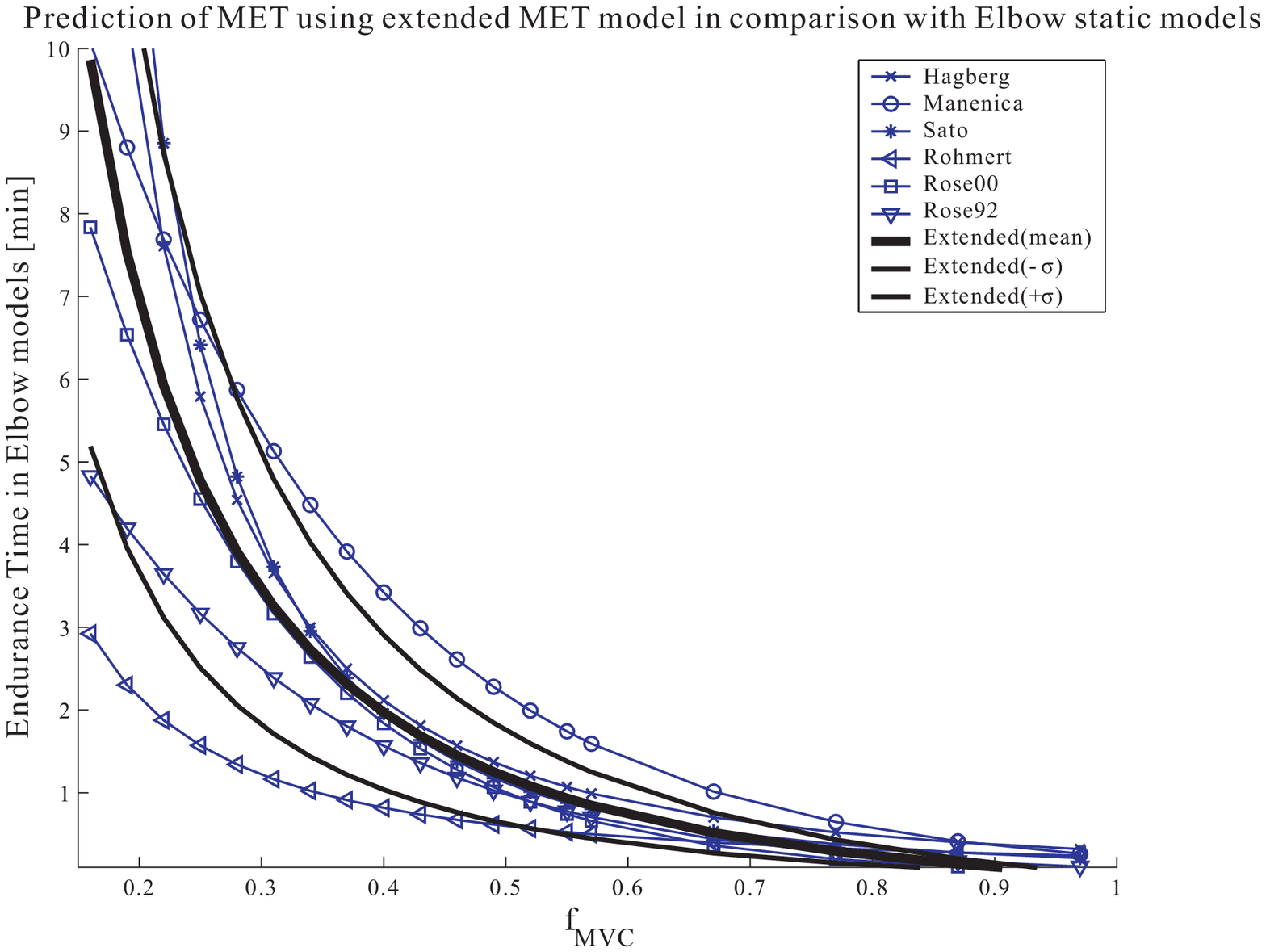}
				\caption{MET prediction using extended MET model in comparison with elbow static models}
				\label{fig:elbowresult}
	    \end{minipage} \\
	    \begin{minipage}[t]{0.45\textwidth}
       	\centering
				\includegraphics[width=1\textwidth]{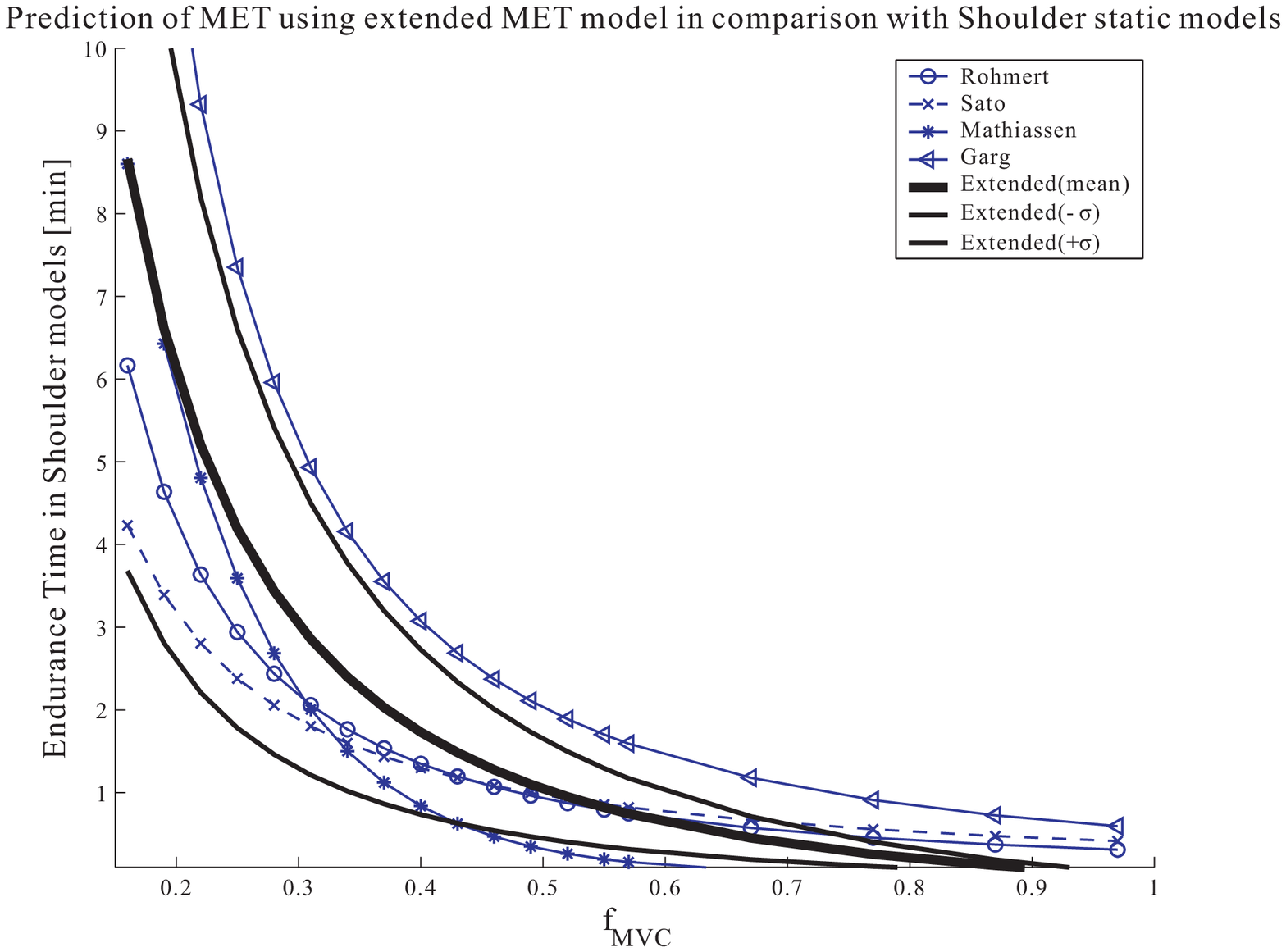}
				\caption{MET prediction using extended MET model in comparison with shoulder static models}
				\label{fig:shoulderresult}
      \end{minipage}&
      \begin{minipage}[t]{0.45\textwidth}
				\centering
				\includegraphics[width=\textwidth]{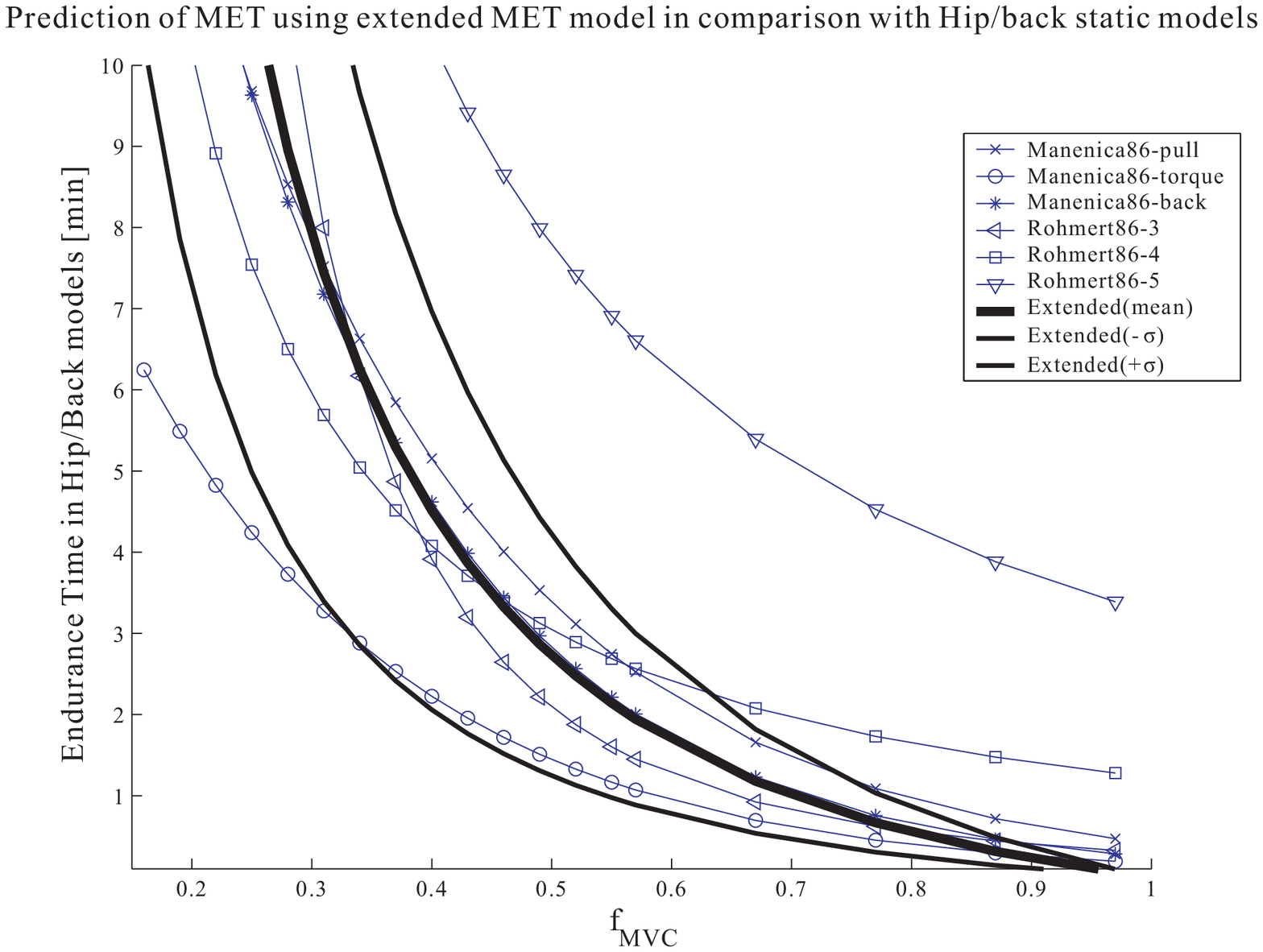}
				\caption{MET prediction using extended MET model in comparison with hip/back static models}
				\label{fig:hipresult}
	    \end{minipage} 

    \end{tabular}
   \end{center}
\end{figure*}

Therefore, the mean value of $\bar{m}$ and its standard deviation $\sigma_m$ are used to redraw the relation between $MET$ and $f_{MVC}$, and they are presented from Fig. \ref{fig:generalresult} to \ref{fig:hipresult}. The black bold solid line is the extended MET model adjusted by $\bar{m}$ and it locates in the range constrained by two slim solid lines adjusted by $\bar{m}\pm\sigma_{m}$. After adjusting our fatigue model with $\bar{m}\pm\sigma_{m}$, the extended MET fatigue model can cover most of the existing MET models from 15\% MVC to 80\% MVC. Although there is an exception in Hip/back model due to the relative large variability in hip muscle groups, it should also be admitted that the adjustment makes the extended MET model suitable for most of the static cases. In another word, the adjustment by mean and deviation makes the extended MET model suitable for evaluating the fatigue for the overall population.

The prediction by the extended MET model cannot cover the models for the \%MVC over 80 as well as the interval under 15\%. However in the industrial cases, it is very rare that the force demand can cross that limit 80\% in static operations. Even if the physical demand beyond 80\%MVC, the prediction difference in the extended MET model from the other MET models is less than one minute.

\subsection{The extended MET model versus previous MET models}
There are several MET models available in the literature, and they cover different body parts.  These models are all experimental models regressed from experimental data, and each model is only suitable for predicting MET of a specific group of people, although the similar tendencies can be found among these models. Furthermore, those MET model cannot reveal individual differences in fatigue characteristic. However, it is admitted that different people might have different fatigue resistances for the same physical operation.

In comparison to conventional MET models, the general analytical MET model was extended from a simple dynamic fatigue model in a theoretical approach. The dynamic muscle fatigue model is based on muscle physiological mechanism. It takes account of task parameters ($F_{load}$ or relative load) and personal factors (MVC and fatigue ratio $k$),  and it  has been validated in comparison to other theoretical models in \citet{ma2008nsd}.  Different from the other MET models, in this extended MET model, there is a parameter $k$ representing individual fatigue characteristic.

After mathematical regression, great similarities ($ICC>0.90$) have been found between the extended MET model and the previous MET models. This indicates that the new theoretical MET model might replace the other MET models by adjusting the parameter $k$. Therefore, the extended MET model generalizes the formation of MET models.

In addition, different fatigue resistances have been found while fitting to different MET models, even for the same muscle group. Therefore, it is interesting to find the influencing factors on the parameter $k$ and to analyze its statistical distribution for ergonomic application. In this paper, we tried to use the mean value and standard deviation of the regressed fatigue resistances. It has been found that the extended MET model with adjustable parameter $k$ could cover most of the MET prediction using experimental MET models. If further experiments can be carried out, it should be promising that the statistical distribution of the fatigue resistance for a given population could be obtained. This kind of information might be useful to integrate early ergonomic analysis into virtual human simulation tools to evaluate fatigue at early work design stage.

\subsection{Influencing factors on fatigue resistance}
Although the MET models fitted from experiment data were formulated in different forms, the $m$ can still provide some useful information for the fatigue resistance, especially for different muscle groups. The differences in fatigue resistance result is possible to be concluded by the mean value and the deviation, but it is still interesting to know why and how the fatigue resistance is different in different muscle groups, in the same muscle group, and even in the same person at different period. There is no doubt that there are several factors influencing on the fatigue resistance of a muscle group, and it should be very useful if the fatigue resistance of different muscle groups can be mathematically modeled. In this section, the fatigue resistance and its variability are going to be discussed in details based on the fatigue resistance results from Table \ref{tab:FatigueResistanceTable} and the previous literature about fatigability. Different influencing factors are going to be discussed and classified in this section. 

All the differences inter muscle groups and intra muscle groups in MET models can be classified into four types: 1) Systematic bias, 2) Fatigue resistance inter individual for constructing a MET model, 3) Fatigue resistance intra muscle group: fatigue resistance differences for the same muscle group, and 4) Fatigue resistance inter muscle groups: fatigue resistance differences for different muscle groups. Those differences can be attributed to different physiological mechanisms involved in different tasks, and influencing variables are subject motivation, central command, intensity and duration of the activity, speed and type of contraction, and intermittent or sustained activities \citep{enoka1995mmf,elfving2007tdb}. In those MET models, all the contractions were exerted under static conditions until exhaustion of muscle groups, therefore, several task related influencing factors can be neglected in the discussion, e.g., speed and duration of contraction. The other influencing factors might contribute to the fatigue resistance difference in MET models.

\textbf{Systematic bias :} all the MET models were regressed or reanalyzed based on experiment results. Due to the experimental background, there were several sources for systematic error. One possible source of the systematic bias comes from experimental methods and model construction \citep{Khalid2006}, especially for the methods with subjective scales to measure MET. The subjective feelings significantly influenced the result. Furthermore, the construction of the MET model might cause system differences for MET model, even in the models which were constructed from the same experiment data (e.g. Huijgens' model and Sjogaard's model in General models). The estimation error was different while using different mathematic models, and it generates systematic bias in the result analysis. 

\textbf{Fatigue resistance inter individual :} besides the systematic error, another possible source for the endurance difference is from individual characteristic. However, the individual characteristic is too complex to be analyzed, and furthermore, the individual characteristic is impossible to be separated from existing MET models, since the MET models already represent the overall performance of the sample participants. In addition, in ergonomic application, the overall performance of a population is often concerned. Therefore, individual fatigue resistance is not discussed in this part separately, but the differences in population in fatigue resistance are going to be discussed and presented in the following part.

\textbf{Fatigue resistance intra muscle group :} the inter individual variability contributes to the errors in constructing MET models and the errors between MET models for the same muscle group. The influencing factors on the fatigue resistance can be mainly classified into sample population characteristic (gender, age, and job), personal muscle fiber composition, and posture.

As mentioned in Section \ref{sec:intro}, the influences on fatigability from gender and age were observed in the literature. In the research for gender influence, women were found with more fatigue resistance than men. Based on muscle physiological principle, four families of factors were adopted to explain the fatigability difference in gender in \citet{hicks2001sdh}. They are: 1) muscle strength (muscle mass) and associated vascular occlusion, 2) substrate utilization, 3) muscle composition and 4) neuromuscular activation patterns. It concluded that although the muscle composition differences between men and women is relatively small \citep{staron2000ftc}, the muscle fiber type area is probably one reason for fatigability difference in gender, since the muscle fiber type I occupied significantly larger area in women than in men \citep{lariviere2006gif}. In spite of muscle fiber composition, the motor unit recruitment pattern acts influences on the fatigability as well. The gender difference in neuromuscular activation pattern was found and discussed in \citet{lariviere2006gif}, and it was observed significantly that females showed more alternating activity between homolateral and contralateral muscles than males. 

Meanwhile, in \citet{mademli2008eva} older men were found with more endurance time then young men in certain fatigue test tasks charging with the same relative load. One of the most common explanations is changes in muscle fiber composition for fatigability change while aging. The shift towards a higher proportion of muscle fiber type I leads old adults having a higher fatigue resistance but smaller MVC. Gender and age were also already taken into a regression model to predict shoulder flexion endurance \citep{Mathiassen99}. In \citet{nussbaum2009effects}, the effects of age, gender, and task parameters on muscle fatigue during prolonged
isokinetic torso exercises were studied. It constated that older men had less initial strength. It was also found that effects of age and gender on fatigue were marginal, while significant interactive effects of age and gender with effort level were found at the same time. 

Besides those two reasons, the muscle fiber composition of muscle varies individually in the population, even in a same age range and in the same gender \citep{staron2000ftc}, and this could cause different performances in endurance tasks. Different physical work history might change the endurance performance. For example, it appeared that athletes with different fiber composition had different advantages in different sports: more type I muscle fiber, better in prolonged endurance events \citep{wilmore2008psa}. Meanwhile, the physical training could also cause shift between different muscle fibers \citep{costill1979asm}. As a result, individual fatigue is very difficult to be determined using MET measurement \citep{Vollestad1997}, and the individual variability might contribute to the differences among MET models for the same muscle group due to selection of subjects.

Back to the existing MET models, the sample population was composed of either a single gender or mixed. At the same time, the number of the subjects was sometimes relative small. For example, only 5 female students (age range 21--33) were measured \citep{Garg2002}, while 40 (20 males, age range 22--48 and 20 females, age range 20--55) were tested in shoulder MET model \citep{Mathiassen99}. Meanwhile, the characteristics of population (e.g., students, experiences workers) could cause some differences in MET studies. Due to different population selection method, different gender composition, and different sample number of participant, fatigue resistance for the same muscle group exists in different experiment results and finally caused different MET models under the similar postures.

In Hip/back models, even with the same sample participants, difference existed also in MET models for different postures. The variation is possible caused by the different MU recruitment strategies and load sharing mechanism under different postures. \citet{kasprisin2000jad} observed that the activation of biceps brachi was significantly affected by joint angle, and furthermore confirmed that joint angle and contraction type contributed to the distinction between the activation of synergistic elbow flexor muscles. The lever of each individual muscle changes along different postures which results different intensity of load for each muscle and then causes different fatigue process for different posture. Meanwhile, the contraction type of each individual muscle might be changed under different posture. Both contraction type change and lever differences contribute to generate different fatigue resistance globally. In addition, the activation difference was also found in antagonist and agonist \citep{karst1987ama, mottram2005mua} muscles as well, and it is implied that in different posture, the engagement of muscles in the action causes different muscle activation strategy, and as a result the same muscle group could have different performances. With these reasons, it is much difficult to indicate the contribution of posture in fatigue resistance because it refers to the sensory-motor mechanism of human, and how the human coordinates the muscles remains not clear enough until yet.

\textbf{Fatigue resistance inter muscle groups:} As stated before, the three different muscle fiber types have different fatigue resistances, and different muscle is composed of types of muscles with composition determining the function of each muscle \citep{Chaffin1999}. The different fatigue resistance can be explained by the muscle fiber composition in different human muscle groups. 

In the literature, muscle fiber composition was used measured by two terms: muscle fiber type percentages and percentage fiber type area (CSA: cross section area). Both terms contribute to the fatigue resistance of the muscle groups. Type I fibers occupied 74\% of muscle fibers in the thoracic muscles, and they amounted 63\% in the deep muscles in lumbar region \citep{sirca1985ftc}. On average type I muscle fibers ranged from 23 to 56\% for the muscles crossing the human shoulder and 12 of the 14 muscles had average SO proportions ranging from 35 to 50\% \citep{dahmane2005sft}. In paper \citep{staron2000ftc,shepstone2005sth}, the muscle fiber composition shows the similar composition for the muscle around elbow and vastus lateralis muscle and the type I muscle fibers have a proportion from 35 - 50\% in average. Although we cannot determine the relationship between the muscle type composition and the fatigue resistance directly and theoretically, the composition distribution among different muscle groups can interpret the MET differences between general, elbow models and back truck models. In addition, the fatigue resistance of older adults is greater than young ones could also be explained by a shift towards a higher proportion of type I fiber composition with aging. These evidences meet the physiological principle of the dynamic muscle fatigue model.

Another possible reason is the loading sharing mechanism of muscles. Hip and back muscle group has the maximum joint moment strength \citep{Chaffin1999} among the important muscle groups. For example, the back extensors are composed of numerous muscle slips having different moment arms and show a particularly high resistance to fatigue relative to other muscle groups \citep{jorgensen1997hte}. This is partly attributed to favorable muscle composition, and the variable loading sharing within back muscle synergists might also contribute significantly to delay muscle fatigue.

In summary, individual characteristics, population characteristics, and posture are external appearance of influencing factors for the fatigue resistance. Muscle fiber composition, muscle fiber area, and sensory motor coordination mechanism are the determinant factors inside the human body deciding the fatigue resistance of muscle group. Therefore, how to construct a bridge to connect the external factors and internal factors is the most important way for modeling the fatigue resistance for different muscle groups. How to combine those factors to model the fatigue resistance remains a challenging work. Despite the difficulty of modeling the fatigue resistance, it is still applicable to find the fatigue resistance for a specified population by MET experiments in regression with the extended MET model due to its simplicity and universal availability. 

\subsection{Limitations}
In the previous discussion, the fatigue resistance of the existing MET models were quantified using $m$ from regression. The possible reasons for the different fatigue resistance were analyzed and discussed. However, how to quantify the influence from different factors on the fatigue resistance remains unknown due to the complexity of muscle physiology and the correlation among different factors. 

The availability of the extended MET model in the interval under 15\% MVC is not validated. The fatigue resistance is only accounted from the 15\% to 99\% MVC due to the unavailability of some MET models under 15\% MVC. For the relative low load, the individual variability under 15\% could be much larger than that over 15\%. The recovery effect might play a much more significant role within such a range.

\section{Conclusions and perspectives}
In this paper, fatigue resistance of different muscle groups were calculated by linear regression from the new fatigue model and the existing MET static models. High $ICC$ has been obtained by regression which proves that our fatigue model can be generalized to predict MET for different muscle groups. Mean and standard deviation in fatigue resistance for different muscle groups were calculated, and it is possible to use both of them together to predict the MET for the overall population. The possible reasons responsible for the variability of fatigue resistance were discussed based on the muscle physiology. 

Our fatigue model is relative simple and computation efficient. With the extended MET model it is possible to carry out the fatigue evaluation in virtual human modeling and ergonomic application, especially for static and quasi-static cases. The fatigue effect of different muscle groups can be evaluated by fitting $k$ from several simple static experiments for certain population. 

\section*{Acknowledgments}
This research was supported by the EADS and the R\'{e}gion des Pays de la Loire (France) in the context of collaboration between the \'{E}cole Centrale de Nantes (Nantes, France) and Tsinghua University (Beijing, PR China).
\bibliographystyle{elsart-harv}

\end{document}